\documentclass[conference,peereview]{IEEEtran}
\IEEEoverridecommandlockouts

\usepackage{hyperref}
\usepackage{xurl} 
\usepackage{cite}
\usepackage{amsmath,amssymb,amsfonts}
\usepackage{algorithmic}
\usepackage[ruled,vlined]{algorithm2e}
\usepackage{graphicx}
\usepackage[table]{xcolor}
\usepackage{tabularx}
\usepackage{multirow}
\usepackage{bm}
\usepackage{forest}
\usepackage{subfigure}
\usepackage[nameinlink,noabbrev]{cleveref}
\usepackage[nolist]{acronym}
\crefformat{equation}{(#2#1#3)}
\Crefformat{equation}{(#2#1#3)}
\usepackage{xcolor,soul}
\usepackage{tikz}
\usetikzlibrary{arrows.meta,positioning,fit}
\usepackage{pgfplots}
\usepgfplotslibrary{groupplots}
\pgfplotsset{compat=1.18}

\usepackage{flushend}

\def\BibTeX{{\rm B\kern-.05em{\sc i\kern-.025em b}\kern-.08em
    T\kern-.1667em\lower.7ex\hbox{E}\kern-.125emX}}

\begin{document}

\title{Detecting Control and Response Events for AI-Enabled Radio Access Networks}

\author{
	\IEEEauthorblockN{Christie Djidjev}
	\IEEEauthorblockA{\textit{Idaho National Laboratory} \\
		Idaho Falls, United States \\
		Christie.Djidjev@inl.gov}
	\and
	\IEEEauthorblockN{Nicholas Kaminski}
	\IEEEauthorblockA{\textit{Idaho National Laboratory} \\
		Idaho Falls, United States\\
		Nicholas.Kaminski@inl.gov}

}

\maketitle

\begin{abstract}
	Next-generation wireless networks are moving toward the use of concurrent AI-driven control functions to optimize different objectives, particularly in AI-RAN and O-RAN architectures. When these functions interact, they can interfere with one another in ways that are difficult to detect from raw network data alone. A key missing piece for managing such interactions is a reliable, interpretable dependency structure that captures which control parameters are actively influencing which network performance outcomes at any given time. This paper focuses on the event-detection step needed to support such dependency learning: given noisy continuous parameter and KPI telemetry, we seek to determine when a genuine control action occurs and when a KPI exhibits a corresponding control-induced response. The difficulty is that KPI fluctuations may also arise from background or exogenous variation, so observed changes cannot be treated directly as control events. To address this challenge, we develop a significance-based event-detection procedure that converts continuous parameter and KPI increments into binary control-activity and KPI-response indicators. To evaluate this procedure, we construct a controlled closed-loop telemetry generator with planted parameter--KPI dependencies and tunable background variation. Experiments show that the proposed procedure reliably detects control-induced events and recovers the underlying dependency structure, outperforming alternative event-detection methods across a range of background-variation levels.
\end{abstract}

\IEEEpeerreviewmaketitle

\begin{IEEEkeywords}
AI-RAN, O-RAN, event detection, Boolean inference, dependency learning, interpretable machine learning, model-based RAN telemetry generation.
\end{IEEEkeywords}

\section{Introduction}
Modern cellular networks increasingly combine independently developed,
artificial intelligence (AI)-enabled control functions for mobility, load balancing, interference
management, slicing, and resource allocation. Open Radio Access Network
(O-RAN) architectures make this control modular through components such as
the Near-Real-Time RAN Intelligent Controller (Near-RT RIC) and
Non-Real-Time RAN Intelligent Controller (Non-RT RIC) \cite{Abdalla2022206,PoleseUnderstandingORAN,ORANSCArchitecture,Montebugnoli20251900}. 
This paper focuses on the AI-for-RAN category of AI-RAN, particularly on applications supporting RAN monitoring and control
\cite{Lindra2025}.
Although each function may behave correctly in isolation, concurrent
functions can interact through shared parameters and overlapping key
performance indicators (KPIs), producing effects that cannot be understood
from any one function alone
\cite{Adamczyk2023199,BrachdelPrever202510590,Giannopoulos2025116684,Zolghadr2025,AlShami2026}.

\begin{figure}[t]
	\centering
	\includegraphics[width=0.7\columnwidth]{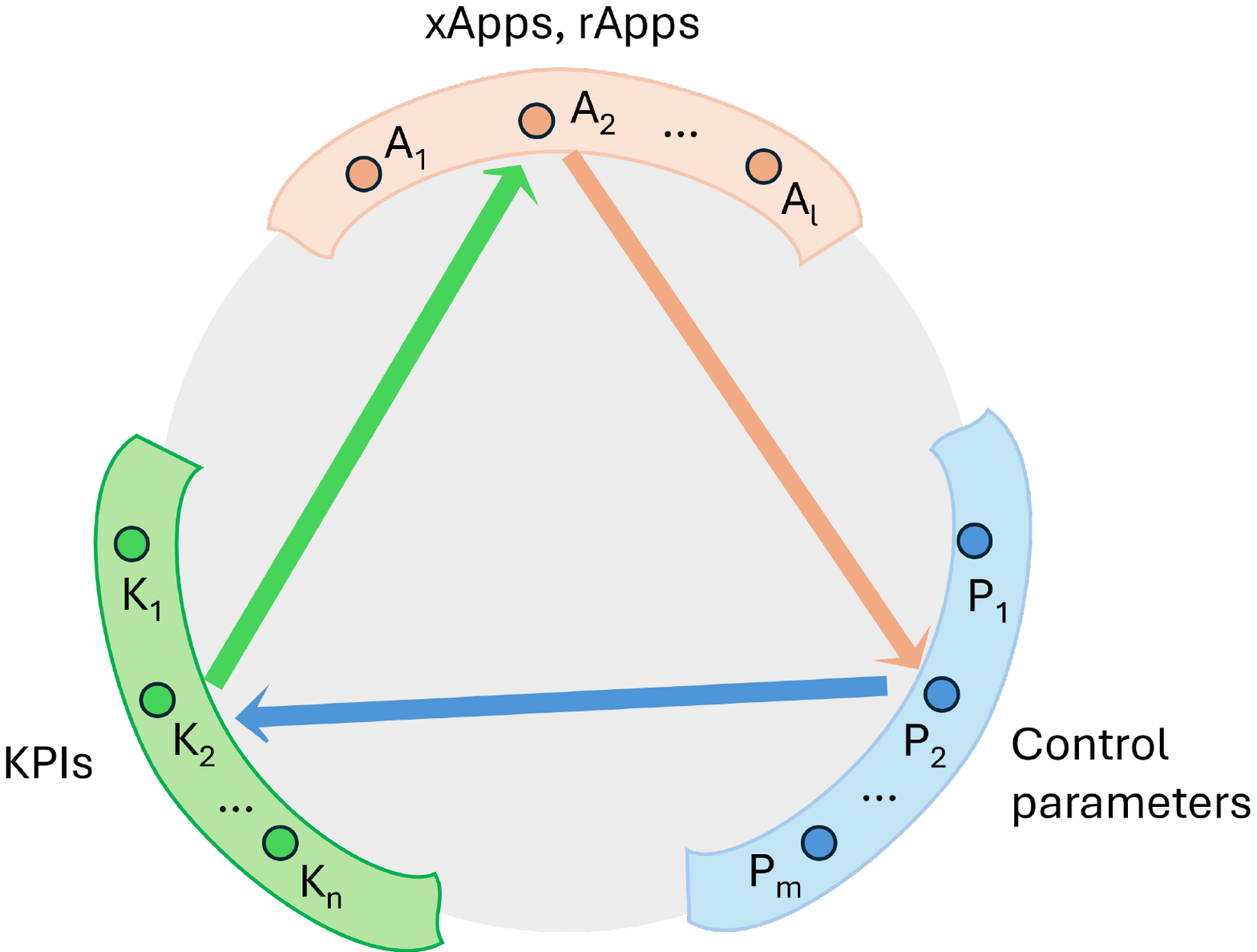}
	\caption{Information flow between control applications, control parameters, and KPIs in AI-RAN/O-RAN. Multiple control functions, such as xApps and rApps, influence network control parameters to meet performance objectives. These parameters, together with external events, influence the observed KPIs. The observed KPIs are then fed back to the applications and influence subsequent control decisions.}
	\label{fig:problem_overview}
\end{figure}

Reasoning about these interactions requires a current, interpretable view of
which control parameters influence which KPIs. As illustrated in
Fig.~\ref{fig:problem_overview}, control applications update parameters,
those updates affect KPIs together with external events, and the observed
KPIs guide later control decisions. Direct conflicts may be identified from
declared parameter use, but indirect and implicit conflicts depend on
parameter--KPI relationships that may be hidden, dynamic, or specific to the
operating context \cite{Zolghadr2025}. A compact dependency representation can support later
conflict analysis, monitoring, and mitigation, but it cannot be obtained by
treating every observed change as evidence of control influence.

Grounded in causal-inference principles that relate interventions to observed
outcomes \cite{peters2017elements}, reasoning about these interactions requires
knowledge of which control actions are actually producing which KPI responses. This paper addresses the first step in building that knowledge from telemetry: detecting actual control actions and corresponding KPI responses in noisy continuous traces.
 Given
continuous parameter increments $\Delta P_i(t)$ and KPI increments
$\Delta K_j(t)$, we encode observed parameter updates as binary activity 
indicators and infer binary indicators of control-induced KPI responses. 
The second task is difficult because background variation can create large KPI changes without a control action,
whereas exogenous variation or overlapping controls can mask a control-induced
response. The resulting event traces are not themselves a learned dependency
matrix; rather, they provide an interpretable intermediate layer from which
dependencies can subsequently be inferred for runtime reasoning
\cite{companion_boolean_paper}.

Evaluation using operational RAN data has so far been limited in the current literature, as no publicly available dataset provides synchronized telemetry with validated event labels and ground-truth parameter–KPI dependencies. Prior O-RAN conflict studies \cite{Wadud2023,Zolghadr2025} use a parameter–KPI model introduced for cognitive autonomous networks in \cite{Banerjee2022}, in which parameter settings are sampled and KPI values are computed from predefined analytic relationships. Such benchmarks are
useful for controlled evaluation, but they do not reproduce the temporal
feedback central to this task or distinguish control-induced responses from
simultaneous exogenous variation. We instead construct closed-loop telemetry
with planted dependencies, overlapping effects, and exact latent event labels,
following the broader use of known data-generating processes in time-series
structure-learning benchmarks \cite{Cheng2024CausalTime}.

We place event extraction in a slow model-refresh loop, leaving lightweight reasoning over its Boolean output to a faster operational loop, such as the runtime reasoning framework described in \cite{companion_boolean_paper}. This division is conceptually consistent with the separation between longer-horizon learning in the Non-RT RIC and near-real-time control in the Near-RT RIC
\cite{Abdalla2022206,PoleseUnderstandingORAN,ORANSCArchitecture}, as well as
two-timescale control principles \cite{Borkar1997,KondaTsitsiklis2004,Kokotovic1999}.

The main contributions of this paper are as follows:
\begin{enumerate}
	\item To the best of our knowledge, we introduce the first event-detection approach specifically designed to identify control actions and control-induced KPI responses from noisy continuous AI-RAN/O-RAN telemetry.
	
	\item We show that our proposed detector advances over standard event-detection baselines, achieving stronger performance than exponentially weighted moving average (EWMA) \cite{lucas1990ewma} and cumulative sum (CUSUM) \cite{page1954cusum} methods in the weak and moderate signal regimes while remaining competitive at high signal strength.
	
	\item We develop a new closed-loop AI-RAN/O-RAN telemetry generator that advances existing model-based generators by providing realistic feedback-driven control dynamics together with exact dependency and event ground truth, enabling systematic evaluation that was not previously possible.
	
\end{enumerate}

\section{Methods}

This section presents the methodology used to study interpretable
parameter--KPI dependency learning. We first construct controlled
AI-RAN-like telemetry with known latent structure, then verify that this
structure is recoverable from the continuous data, and finally address
the main methodological problem of detecting binary control-activity and
KPI-response events from continuous telemetry. These events form an
intermediate representation intended to support subsequent
parameter--KPI dependency learning.

\begin{table*}[t]
	
	\caption{Exhaustive Decision Cases in the Closed-Loop Telemetry Generator}
	\label{tab:generator_logic}
	
	\centering
	\scriptsize
	
	\renewcommand{\arraystretch}{1.15}
	\setlength{\tabcolsep}{3.5pt}
	
	\begin{tabularx}{\textwidth}{
			|c|c|c|c|c|c|c|c|X|}
		
		\hline
		
		\shortstack{\rule{0pt}{2.2ex}%
			$\bm{K}_j$ \textbf{in}\\
			\textbf{preferred}\\
			\textbf{range?}
		}
		&
		$\bm{L}^{\ast}_{j,i}$
		&
		\shortstack{
			\textbf{Other } $\bm{K}_{j'}$\\
			\textbf{activates } $\bm{P}_i$\textbf{?}\\[-0.5mm]
			$j'\neq j$
		}
		&
		\shortstack{
			\textbf{Other } $\bm{P}_{i'}$\\
			\textbf{active on } $\bm{K}_j$\textbf{?}\\[-0.5mm]
			$i'\neq i$
		}
		&
		\shortstack{
			$\bm{|\Delta P_i|}$\\
			$\bm{> \tau_{P_i}}$\textbf{?}
		}
		&
		$\bm{\Delta K_j}$
		&
		$\bm{B}_P^{\mathrm{GT}}$
		&
		$\bm{B}_K^{\mathrm{GT}}$
		&
		\textbf{Note}
		\\
		
		\hline
		
		Any & 0 & No & No &
		No &
		$\displaystyle E$ &
		0 & 0 &
		No control contribution from $P_i$ or other parameters
		\\
		
		\hline
		
		Any & 0 & Yes & No &
		Yes &
		$\displaystyle E$ &
		1 & 0 &
		$P_i$ changes but does not affect $K_j$
		\\
		
		\hline
		
		Any & 0 & No & Yes &
		No &
		$\displaystyle \sum_{i'\neq i} C_{i'} + E$ &
		0 & 1 &
		Another parameter provides the control response of $K_j$
		\\
		
		\hline
		
		Any & 0 & Yes & Yes &
		Yes &
		$\displaystyle \sum_{i'\neq i} C_{i'} + E$ &
		1 & 1 &
		$P_i$ changes independently of the control response of $K_j$
		\\
		
		\hline
		
		Yes & 1 & No & No &
		No &
		$\displaystyle E$ &
		0 & 0 &
		Planted edge inactive at this timestep
		\\
		
		\hline
		
		Yes & 1 & Yes & No &
		Yes &
		$\displaystyle C_i + E$ &
		1 & 1 &
		Another KPI activates $P_i$, which also affects $K_j$
		\\
		
		\hline
		
		Yes & 1 & No & Yes &
		No &
		$\displaystyle \sum_{i'\neq i} C_{i'} + E$ &
		0 & 1 &
		$P_i$ is inactive, but other controls affect $K_j$
		\\
		
		\hline
		
		Yes & 1 & Yes & Yes &
		Yes &
		$\displaystyle C_i + \sum_{i'\neq i} C_{i'} + E$ &
		1 & 1 &
		Overlapping control effects on $K_j$
		\\
		
		\hline
		
		No & 1 & Any & No &
		Yes &
		$\displaystyle C_i + E$ &
		1 & 1 &
		Corrective action; response may be masked if
		$\displaystyle C_i + E \approx 0$
		\\
		
		\hline
		
		No & 1 & Any & Yes &
		Yes &
		$\displaystyle C_i + \sum_{i'\neq i} C_{i'} + E$ &
		1 & 1 &
		Corrective action with overlapping controls
		\\
		
		\hline
		
	\end{tabularx}
	
	\vspace{2mm}
	\begin{minipage}{0.97\textwidth}
		\footnotesize
		\textit{Notes:}
		\(L^\ast_{j,i}=1\) indicates a planted dependency
		\(P_i\!\rightarrow K_j\).
		Parameter activity is represented by
		\(B_P^{\mathrm{GT}}(i,t)
		=\mathbf{1}\{|\Delta P_i(t)|>\tau_{P_i}\}\);
		in the present implementation, \(\tau_{P_i}=0\), so this indicator
		records whether a control-driven update of \(P_i\) occurs.
		\(B_K^{\mathrm{GT}}(j,t)=1\) indicates that \(K_j\) has a
		control-induced contribution at that timestep, irrespective of
		whether exogenous variation masks or amplifies that contribution
		in the observed \(\Delta K_j\).
		\(C_i\) denotes the contribution to \(\Delta K_j\) from an active
		parameter \(P_i\), while \(E\) denotes exogenous KPI variation.
		``Any'' means that the corresponding input does not affect the
		outcome for that row.
	\end{minipage}
\end{table*}

\subsection{Data Generation}

The main methodological focus of this paper is the detection of
control-activity and KPI-response events from continuous telemetry, as
described later in this section. Evaluating such a procedure first
requires telemetry for which the underlying control--KPI relationships
and the true event states are known. This creates a fundamental data
availability problem.
No publicly available AI-RAN/O-RAN dataset provides synchronized control and KPI traces together with validated
ground-truth parameter--KPI dependencies.
Existing model-based O-RAN benchmarks sample parameter settings independently
and compute KPI values from fixed analytic relationships
\cite{Banerjee2022,Wadud2023,Zolghadr2025}. Consequently, they do not capture
feedback-driven temporal control behavior or distinguish control-induced KPI
responses from exogenous variation and overlapping control effects, which are
the distinctions required for the event-detection task considered here. We
therefore develop a closed-loop RAN-inspired generator with known dependencies
and event labels.

Specifically, we model \(n_P\) control parameters \(P_i(t)\),
\(i=1,\ldots,n_P\), and \(n_K\) KPIs \(K_j(t)\),
\(j=1,\ldots,n_K\), observed over discrete timesteps. The generator
represents a broad class of feedback-driven control functions in which
xApps or rApps observe KPI values and modify control parameters in
response to deviations from desired operating regions. It is not
intended as a universal model of all xApps or rApps, some of which may
use longer histories, optimization over extended horizons, or learned
policies. Rather, it provides the controlled closed-loop setting needed
for the event-detection experiments in this paper.

The latent parameter--KPI structure is encoded by a binary matrix
\(L^\ast \in \{0,1\}^{n_K\times n_P}\), where
\(L^\ast_{j,i}=1\) means that parameter \(P_i\) can affect KPI \(K_j\),
and \(L^\ast_{j,i}=0\) means that it cannot. We also associate a sign
matrix \(S\in\{-1,0,1\}^{n_K\times n_P}\) with these dependencies,
where \(S_{j,i}\) indicates whether increasing \(P_i\) tends to move
\(K_j\) upward or downward whenever \(L^\ast_{j,i}=1\). Together,
\(L^\ast\) and \(S\) define the hidden control structure that the
learning procedure is ultimately expected to recover.

Each KPI \(K_j\) is assigned a preferred operating interval
\([K_{j,\mathrm{pref}}^{\min},K_{j,\mathrm{pref}}^{\max}]\), which functions
as a deadband in the feedback loop: while \(K_j\) remains inside this interval,
it does not trigger corrective control \cite{astrom2008feedback}. When it moves
outside the preferred interval, the corresponding controller may update
parameters connected to that KPI.
Once a parameter changes, however, its effect is not restricted to the
KPI that triggered the update: the change contributes to every KPI to
which that parameter is connected through \(L^\ast\). Consequently,
one control action may affect several KPIs, and several simultaneously
active parameters may contribute to the response of the same KPI.

In addition to these control-induced effects, each KPI is subject to
exogenous variation. For each KPI and timestep, the generator draws an
independent zero-mean Gaussian contribution
\[
E_j(t)\sim\mathcal{N}(0,\sigma^2),
\]
where \(\sigma\) controls the background-variation scale. The observed
KPI change therefore combines the contributions of all active connected
parameters with \(E_j(t)\). Parameter and KPI values are constrained to
their configured operating ranges.

The generator also retains the latent state from which each transition
is produced. We denote by
\(B_P^{\mathrm{GT}}(i,t)\in\{0,1\}\) the ground-truth indicator that a
control-driven update of parameter \(P_i\) occurs at time \(t\), and by
\(B_K^{\mathrm{GT}}(j,t)\in\{0,1\}\) the ground-truth indicator that
there is a control-induced contribution to the response of \(K_j\) at
that timestep. Importantly, \(B_K^{\mathrm{GT}}(j,t)=1\) refers to the
latent control contribution, regardless of whether exogenous variation
amplifies, reduces, or nearly cancels that contribution in the observed
\(\Delta K_j(t)\). These labels, together with \(L^\ast\), are available
only because the data are generated in a controlled setting and are not
provided to the event-detection procedure.

The decision logic is detailed in Table~\ref{tab:generator_logic}. The first four columns describe the control state
relevant to a parameter--KPI pair \((P_i,K_j)\), while the remaining
columns show the resulting parameter activity, observed KPI increment,
and ground-truth event states.

Table~\ref{tab:generator_logic} highlights why the event-detection problem is nontrivial even in this controlled setting. A parameter may change because of a KPI other than
the \(K_j\) under consideration, and \(K_j\) may simultaneously receive
contributions from several active parameters. Moreover, exogenous
variation may produce a substantial observed \(\Delta K_j\) when no
control-induced response is present, or may partially cancel a genuine
control response and make the resulting observed change small.
Consequently, neither parameter activity nor the magnitude of an
observed KPI increment alone uniquely determines the underlying
parameter--KPI event state.

To reduce dependence on any single initial network state, a generated
dataset may contain \(n_{\mathrm{start}}\) separate closed-loop
trajectories. For each trajectory, the initial parameter and KPI values
are sampled independently and uniformly from their configured ranges.
The requested number of timesteps is divided approximately evenly among
the starting points, and the resulting trajectories are concatenated
to form the dataset.

The main generator settings are therefore the dependency matrix
\(L^\ast\), the sign matrix \(S\), the intervention step size
\(\Delta P_{\mathrm{step}}\), the preferred KPI intervals
\(\{[K_{j,\mathrm{pref}}^{\min},
K_{j,\mathrm{pref}}^{\max}]\}_{j=1}^{n_K}\), the configured parameter
and KPI ranges, the total trajectory length, the exogenous-variation
scale \(\sigma\), and \(n_{\mathrm{start}}\). Structural properties such as sparsity and overlap are determined by
\(L^\ast\). Here, sparsity describes how few parameter--KPI relationships
are present, while overlap describes how often multiple parameters affect
the same KPI or one parameter affects multiple KPIs. Of
particular importance for the event-detection stage is the balance
between the magnitude of the control-induced KPI response and
\(\sigma\), since this determines how clearly genuine responses can be
distinguished from background variation.

The generator records both absolute trajectories
\((P(t),K(t))\) and incremental traces
\((\Delta P(t),\Delta K(t))\). The incremental traces are supplied to
the event-detection procedure described below. That procedure does not
have access to \(L^\ast\), the controller's internal decision state, or
the ground-truth KPI-response labels \(B_K^{\mathrm{GT}}\).

\subsection{Recoverability from Continuous Traces}

Before evaluating the event detector, we first verify that the planted
dependency structure is recoverable from the underlying continuous
telemetry. This is not the primary inference task studied in this paper;
rather, it establishes that the generated traces contain the dependency
structure before they are reduced to binary events. Because event
detection replaces the continuous traces with a coarser representation,
failure to recover \(L^\ast\) from the continuous data would make
subsequent recovery from the binary representation difficult to
interpret. Continuous-trace recoverability therefore serves as a
verification of the benchmark before the event-detection stage.

To carry out this check, we use decision trees as predictive models, chosen for their interpretability in the continuous setting. For each KPI \(K_j\), we define a supervised prediction problem in which the input at time \(t\) is
\begin{align}
	x(t)=
	\Delta P_1(t),\ldots,\Delta P_{n_P}(t),\nonumber ~~~~~~~~~~~~~~~~~~~~~ \\ 
	P_1(t),\ldots,P_{n_P}(t),K_1(t),\ldots,K_{n_K}(t),
\end{align}
\noindent
and the target is
\[
y_j(t)=\Delta K_j(t).
\]

To train the tree as an interpretable classifier, we replace the continuous target \(y_j(t)\) with a discretized label
\[
\tilde{y}_j(t)=Q(\Delta K_j(t)),
\]
where \(Q(\cdot)\) maps each KPI increment to the nearest value in the discrete set
\[
V=\{-2\Delta P_{\mathrm{step}},-\Delta P_{\mathrm{step}},0,\Delta P_{\mathrm{step}},2\Delta P_{\mathrm{step}}\}.
\]
Thus, \(\tilde{y}_j(t)\) records which canonical signed increment is closest to the observed \(\Delta K_j(t)\). In the experiments reported here, \(\Delta P_{\mathrm{step}}=5\), so the discrete target values are \(\{-10,-5,0,5,10\}\). For readability in the tree diagrams, we denote these five classes symbolically by
\(
\{--,\,-,\,=,\,+,\,++\},
\)
corresponding respectively to \(\{-10,-5,0,5,10\}\). This preserves more information than a pure change/no-change label while still allowing the use of an interpretable classifier.

The verification procedure has three main stages. First, for each KPI \(K_j\), we train a decision tree to predict its discretized increment from the current system state and the current parameter changes. Second, for each pair \((K_j,P_i)\), we inspect the learned tree for \(K_j\) and count how many nonredundant splits occur on the feature \(\Delta P_i\); we denote this count by \(N(K_j,P_i)\). Third, we aggregate these counts across all KPI--parameter pairs to construct an estimated binary dependency matrix \(L_{\mathrm{pred}}\).

\subsubsection{Decision-tree prediction of KPI increments}

For each KPI \(K_j\), we train a separate decision-tree model using the input--target pairs \((x(t), \tilde{y}_j(t))\) defined above. Including both \(\Delta P\) and the absolute state variables \((P,K)\) serves a specific purpose. The \(\Delta P\) features describe what control action was taken at the current step, while the \(P\) and \(K\) features describe the operating regime in which that action occurred. Together, these features allow the decision tree to model KPI dynamics from the continuous state rather than trying to infer \(L^\ast\) from parameter changes alone.

\subsubsection{Counting nonredundant parameter splits}

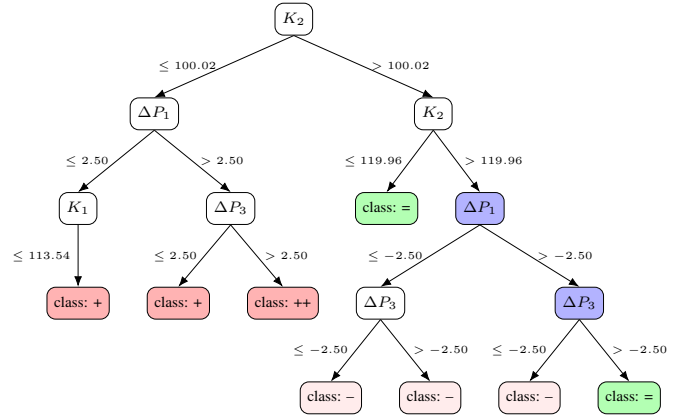
\begin{figure}[t]
	\centering
	\resizebox{\columnwidth}{!}{%
		\begin{forest}
			for tree={
				draw,
				rounded corners,
				align=center,
				edge={-Latex},
				parent anchor=south,
				child anchor=north,
				font=\scriptsize,
				l sep=10mm,
				s sep=6mm,
			}
			[
			{$K_2$}
			[
			{$\Delta P_1$},
			edge label={node[midway,left] {\tiny $\le 100.02$}}
			[
			{$K_1$},
			edge label={node[midway,left] {\tiny $\le 2.50$}}
			[
			{class: +},
			fill=red!30,
			edge label={node[midway,left] {\tiny $\le 113.54$}}
			]
			]
			[
			{$\Delta P_3$},
			edge label={node[midway,right] {\tiny $> 2.50$}}
			[
			{class: +},
			fill=red!30,
			edge label={node[midway,left] {\tiny $\le 2.50$}}
			]
			[
			{class: ++},
			fill=red!30,
			edge label={node[midway,right] {\tiny $> 2.50$}}
			]
			]
			]
			[
			{$K_2$},
			edge label={node[midway,right] {\tiny $> 100.02$}}
			[
			{class: =},
			fill=green!30,
			edge label={node[midway,left] {\tiny $\le 119.96$}}
			]
			[
			{$\Delta P_1$},
			fill=blue!30,
			edge label={node[midway,right] {\tiny $> 119.96$}}
			[
			{$\Delta P_3$},
			edge label={node[midway,left] {\tiny $\le -2.50$}}
			[
			{class: --},
			fill=pink!30,
			edge label={node[midway,left] {\tiny $\le -2.50$}}
			]
			[
			{class: --},
			fill=pink!30,
			edge label={node[midway,right] {\tiny $> -2.50$}}
			]
			]
			[
			{$\Delta P_3$},
			fill=blue!30,
			edge label={node[midway,right] {\tiny $> -2.50$}}
			[
			{class: --},
			fill=pink!30,
			edge label={node[midway,left] {\tiny $\le -2.50$}}
			]
			[
			{class: =},
			fill=green!30,
			edge label={node[midway,right] {\tiny $> -2.50$}}
			]
			]
			]
			]
			];
		\end{forest}%
	}
	\caption{Example learned decision tree for \(\Delta K_2\). Leaves output discretized classes, shown symbolically as \(\{--,-,=,+,++\}\), corresponding respectively to the target values \(\{-10,-5,0,5,10\}\). Leaf colors distinguish prediction classes, while blue internal nodes mark parameter-increment splits that are treated as evidence of dependency because their branches can lead to different downstream predicted classes.}
	\label{fig:dt_k2}
\end{figure}

After one decision tree has been trained for each KPI, the second stage inspects the learned trees to quantify evidence of dependency. The key idea is that a parameter should be counted as influential for KPI \(K_j\) only if the tree for \(K_j\) uses the corresponding increment feature \(\Delta P_i\) in a way that actually changes the model's downstream prediction. A mere appearance of \(\Delta P_i\) in the tree is not enough, since a tree may contain a split that partitions the data without changing the predicted outcome. Figure~\ref{fig:dt_k2} illustrates this criterion.

Accordingly, for a fixed KPI \(K_j\) and parameter \(P_i\), we inspect all tree nodes that split on \(\Delta P_i\). A split is counted as nonredundant if its two child subtrees can lead to different downstream predictions for \(\Delta K_j\). We denote by \(N(K_j,P_i)\) the number of such nonredundant splits in the tree trained for KPI \(K_j\).

\subsubsection{Stage 3: Constructing the estimated support matrix}

In the third stage, we convert the counts \(N(K_j,P_i)\) into an estimated binary dependency matrix \(L_{\mathrm{pred}}\). For each KPI--parameter pair \((K_j,P_i)\), we declare an influence to be present if the corresponding tree contains at least one nonredundant split on \(\Delta P_i\). Formally, we define
\[
L_{\mathrm{pred}}[j,i]
=
\begin{cases}
	1, & \text{if } N(K_j,P_i) \ge 1,\\
	0, & \text{otherwise.}
\end{cases}
\]

This aggregation yields an interpretable recoverability test: \(L_{\mathrm{pred}}[j,i]=1\) if and only if the tree for \(K_j\) contains at least one nonredundant split on \(\Delta P_i\).

\subsection{Control--KPI Event Detection from Continuous Telemetry}

Given the observed continuous increments \(\Delta P_i(t)\) and
\(\Delta K_j(t)\), the objective is to determine when a genuine control
action occurs and when a KPI exhibits a control-induced response.
Observed KPI changes cannot be interpreted directly as such events:
background variation may produce a KPI change in the absence of a
control response, while a genuine response may be amplified or partially
masked by exogenous variation or simultaneous control effects. We
therefore seek a binary event representation that preserves meaningful
control activity and KPI responses while suppressing changes attributable
to the background process.

This task differs from the recoverability analysis above. There,
continuous telemetry and an interpretable predictive model were used
only to verify that the planted dependency structure is present in the
generated data. Here, the task is to infer binary event indicators
directly from the observed parameter and KPI increments. These event traces are intended as input to a downstream binary support-inference procedure, which is outside the present evaluation.


The Booleanization stage is deliberately given only the observed
continuous increments \(\Delta P_i(t)\) and \(\Delta K_j(t)\), and not
the generator's latent dependency structure, control decisions, or
ground-truth event labels summarized in
Table~\ref{tab:generator_logic}.  In particular, a nonzero or even large
KPI increment does not necessarily correspond to a true control-induced
response, while a true response may be partially obscured by background
variation or simultaneous control effects.  The task is therefore to
infer the underlying event state from the continuous telemetry rather
than to reproduce a Boolean state directly encoded in the data.

We now Booleanize both the parameter and KPI increments. On the parameter side, the goal is to construct a binary activity indicator
\[
B_P(i,t)\in\{0,1\}
\]
that records whether parameter \(P_i\) underwent a meaningful update at timestep \(t\). In the current generator, parameter updates are discrete control actions rather than noisy measurements, so we do not introduce a separate noise term on the \(P\)-side. Accordingly, the present implementation defines
\[
B_P(i,t)=
\begin{cases}
	1, & \text{if } \Delta P_i(t)\neq 0,\\
	0, & \text{if } \Delta P_i(t)=0.
\end{cases}
\]
Thus, a null row is a timestep at which \(B_P(i,t)=0\) for all \(i\), equivalently, a timestep at which all parameter increments are zero. Because parameter updates are directly observed and are not corrupted by
measurement noise in the present generator, this indicator coincides
with the corresponding ground-truth state
\(B_P^{\mathrm{GT}}(i,t)\); the more challenging inference problem is on
the KPI side.

On the KPI side, the goal is to construct a binary response indicator
\[
B_K(j,t)\in\{0,1\}
\]
that records whether KPI \(K_j\) exhibits a significant response at timestep \(t\). Unlike the parameter side, this is not immediate, because even when no control action occurs, \(\Delta K_j(t)\) may still vary because of external events. We therefore estimate a KPI-specific no-update baseline from null rows, i.e., timesteps at which no parameter update occurs.  Formally, let
\[
\mathcal{T}_0=\left\{t:\sum_{i=1}^{n_P} |\Delta P_i(t)| = 0\right\}
\]
denote the set of null rows. 
For each KPI \(K_j\), we collect the null-row increments
\[
\{\Delta K_j(t): t\in\mathcal{T}_0\},
\]
and estimate the null baseline parameters by
\[
\hat{\mu}_j = \frac{1}{|\mathcal{T}_0|}\sum_{t\in\mathcal{T}_0}\Delta K_j(t),
\]
\[
\hat{\sigma}_j =
\max\!\left\{
\sqrt{
	\frac{1}{|\mathcal{T}_0|-1}
	\sum_{t\in\mathcal{T}_0}\bigl(\Delta K_j(t)-\hat{\mu}_j\bigr)^2
},
\,10^{-8}
\right\}.
\]
We then standardize each KPI increment as
\[
z_j(t)=\frac{\Delta K_j(t)-\hat{\mu}_j}{\hat{\sigma}_j},
\]
and declare a significant KPI response when
\begin{equation}
B_K(j,t)=\mathbf{1}\{|z_j(t)|>z_{\mathrm{th}}\},\label{eq:zth}
\end{equation}
where \(z_{\mathrm{th}}\) is a tunable significance threshold. The
threshold values examined in the experiments were chosen to span a broad
range from permissive to highly selective detection rather than to optimize
performance for a particular dependency graph. Since \(z_j(t)\) is
standardized with respect to the control-inactive baseline, increasing
\(z_{\mathrm{th}}\) has the direct interpretation of requiring a KPI change
to lie progressively farther into the tail of its null distribution.
Accordingly, the threshold sweep in Fig.~\ref{fig:detection_quality_main}(b)
is used to expose the precision--recall tradeoff and identify the regime in
which performance stabilizes. In an operational setting,
\(z_{\mathrm{th}}\) need not be fixed a priori: it can instead be calibrated
from held-out control-inactive samples to achieve a desired false-alarm
rate. We use this latter procedure in the baseline comparison, where all
detectors are calibrated to the same target null false-alarm rate of
\(10^{-3}\).

Algorithmically, the procedure for the KPIs has three steps: first identify the null rows \(\mathcal{T}_0\) from timesteps with no parameter update; second estimate \(\hat{\mu}_j\) and \(\hat{\sigma}_j\) for each KPI from the corresponding null-row increments; third compute the z-scores \(z_j(t)\) and threshold them to obtain \(B_K(j,t)\). Together with the binary parameter-activity indicators \(B_P(i,t)\), these binary KPI-response indicators form an event representation intended for downstream dependency inference.

\section{Experimental Results and Discussion}

This section evaluates the main stages of the proposed learning-loop pipeline. We begin by verifying that the latent dependency structure is recoverable from the synthetic continuous traces before any Booleanization is applied. We then examine the Booleanization step itself, focusing on how signal strength, detector threshold, and latent dependency size affect detection quality. Finally, we report the runtime of the main offline stages to assess whether the interpretable model-refresh pipeline remains practical at the slower timescale for which it is intended.

\begin{table}[t]
	\centering
	\caption{Mean recoverability of the latent dependency structure from continuous traces before Booleanization.}
	\label{tab:recoverability}
	\setlength{\tabcolsep}{1pt}
	\begin{tabular}{|l|c|c|c|c|}
		\hline
		Condition & ~~Precision~~ & ~~Recall~~ & ~~~~F1~~~~ & Exact-match accuracy \\
		\hline
		Overall          & 0.980 & 0.975 & 0.975 & 0.946 \\
		Sparse           & 0.958 & 0.958 & 0.958 & 0.980 \\
		Moderate overlap~ & 0.987 & 0.983 & 0.982 & 0.961 \\
		High overlap     & 0.983 & 0.973 & 0.975 & 0.908 \\
		\hline
	\end{tabular}
\end{table}

\subsection{Recoverability from continuous traces}

Before studying Booleanization, we first ask a simpler question: is the dependency structure visible in the continuous traces at all? To check this, we train one decision tree per KPI to predict discretized \(\Delta K_j\) from \((\Delta P, P, K)\), and then convert the learned trees into an estimated influence matrix \(L_{\mathrm{pred}}\). This serves as a sanity check on the generator: if a simple interpretable model can recover \(L^\ast\), then the synthetic traces contain the structure we want to study.

Table~\ref{tab:recoverability} summarizes this continuous-trace recoverability step. Recoverability is strong overall, with micro-averaged precision, recall, and F1 all close to \(0.98\), and an exact-match accuracy of \(0.946\). This indicates that the latent dependency structure is already well represented in the continuous traces before Booleanization is applied.

The main trend is that overlap affects full-matrix recovery more clearly than edge-wise recovery. In the high-overlap regime, precision, recall, and F1 remain high, but exact-match accuracy drops to \(0.908\), indicating that denser dependency patterns are more likely to incur at least one structural mistake even when most individual edges are recovered correctly. By contrast, the sparse regime achieves the highest exact-match accuracy, although its edge-level metrics show larger variability across cases. Overall, these results support the intended role of this step: the continuous data preserve the latent support pattern well enough for a simple interpretable model to recover it, so later errors are more naturally attributed to the Booleanization stage rather than to the generator hiding the structure too well.

\subsection{Detection quality under signal strength, threshold choice, and dependency size}

\begin{figure}[t]
\centering
\includegraphics[width=\columnwidth]{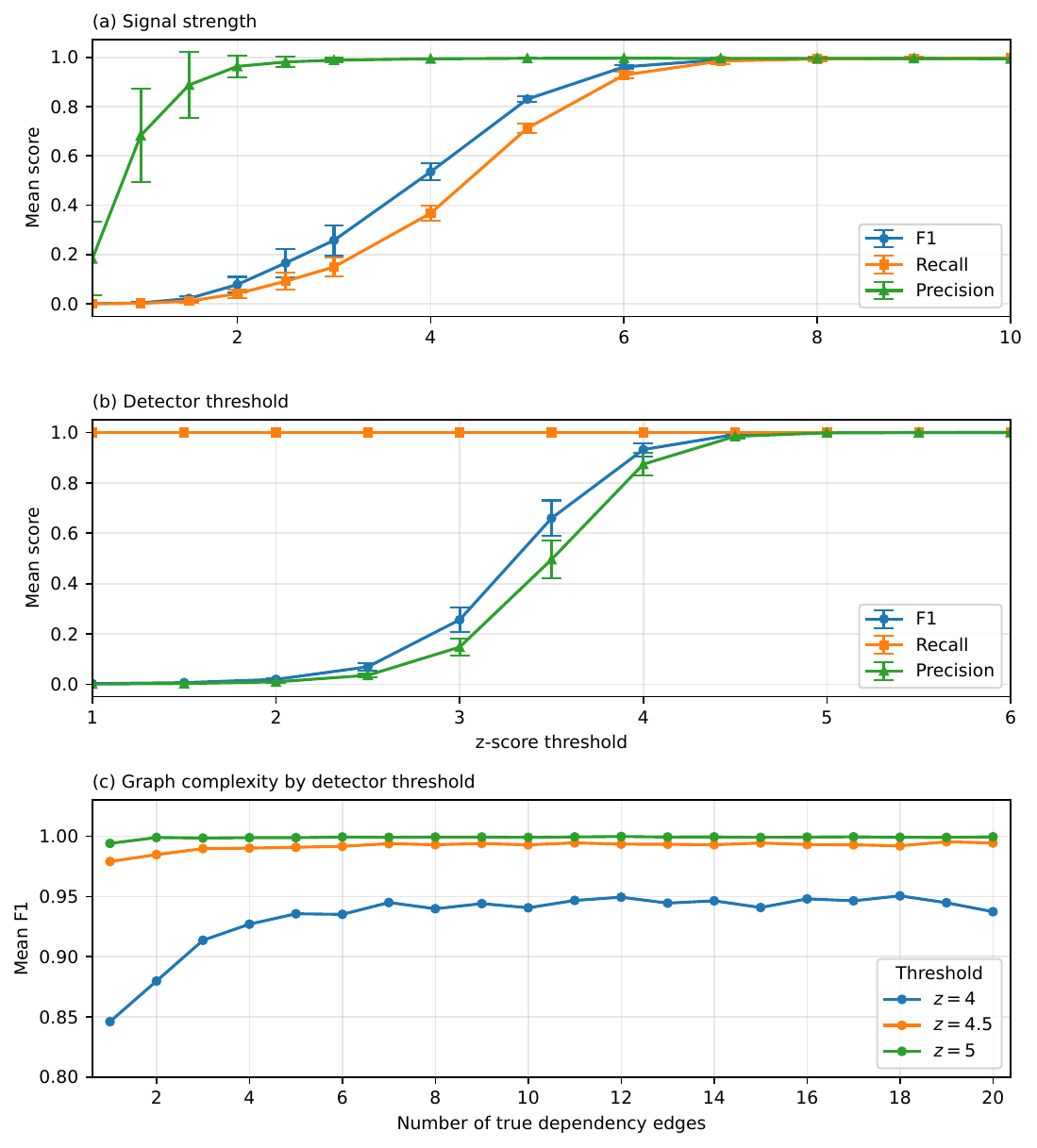}
\caption{Detection-quality trends for the binary event-detection step. (a) Mean F1, recall, and precision versus signal strength, measured by $\Delta P_{\mathrm{step}}/\sigma$. (b) Mean F1, recall, and precision versus detector threshold $z_{\mathrm{th}}$. (c) Mean F1 versus the number of true parameter--KPI dependency edges for selected detector thresholds.}
\label{fig:detection_quality_main}
\end{figure}

Fig.~\ref{fig:detection_quality_main} summarizes the Booleanization step from three complementary viewpoints. Panel~(a) studies signal strength, measured by $\Delta P_{\mathrm{step}}/\sigma$, which controls how clearly control-induced KPI responses separate from background variation. Panel~(b) studies the detector threshold $z_{\mathrm{th}}$ used in \cref{eq:zth} to declare a KPI-response event. Panel~(c) shows how the same threshold tradeoff behaves across latent dependency structures of different sizes, measured by the number of true parameter--KPI edges in the planted graph. Taken together, these panels distinguish limits imposed by the data itself from limits imposed by detector calibration or by structural complexity.

Panel~(a) shows a clear weak-signal to strong-signal transition. At small values of $\Delta P_{\mathrm{step}}/\sigma$, recall is poor because true KPI responses are difficult to separate from exogenous variation. As signal strength increases, recall rises sharply and F1 improves with it, while precision is already high and quickly saturates. This indicates that the main challenge in the low signal-to-background-ratio regime is missed detections rather than a large false-positive burden.

Panel~(b) shows that threshold calibration provides a transparent operating-point tradeoff. At low thresholds, recall remains high, but precision is poor, which indicates that many background fluctuations are labeled as signal. As the threshold increases, precision rises sharply while recall stays high over much of the range, and F1 correspondingly improves before saturating. This suggests that the Booleanization step is most effective in a moderate-to-high threshold regime, where the detector has become selective without sacrificing much sensitivity.

Panel~(c) examines true dependency edges more directly by plotting mean F1
as a function of the number of true parameter--KPI dependency edges for
\(z_{\mathrm{th}}\in\{4,4.5,5\}\). These values were selected from the
high-selectivity region identified by the broader threshold sweep in
Panel~(b), rather than tuned separately for individual dependency graphs.
The comparison therefore tests whether detector performance remains stable
with increasing graph complexity once a selective operating regime has
been reached. Overall, these results suggest that once the detector threshold is sufficiently selective, Boolean event-detection quality is largely stable across the tested range of dependency-graph sizes, with only modest variation at the lowest threshold shown.

This performance is notable in light of
Table~\ref{tab:generator_logic}: the detector uses only a KPI-specific
null baseline and a single significance threshold, without access to the
planted dependency graph or the ground-truth control state, yet it
recovers the underlying KPI-event labels accurately over a broad range
of the tested conditions.

Overall, these results show that Booleanization quality is governed first by the separation between control-induced responses and the background variation floor, and second by calibration of the detector threshold. Once the signal is sufficiently strong, the remaining task is mainly to choose a threshold that suppresses spurious detections without discarding true events.

\begin{figure}[t]
\centering
\includegraphics[width=\columnwidth]{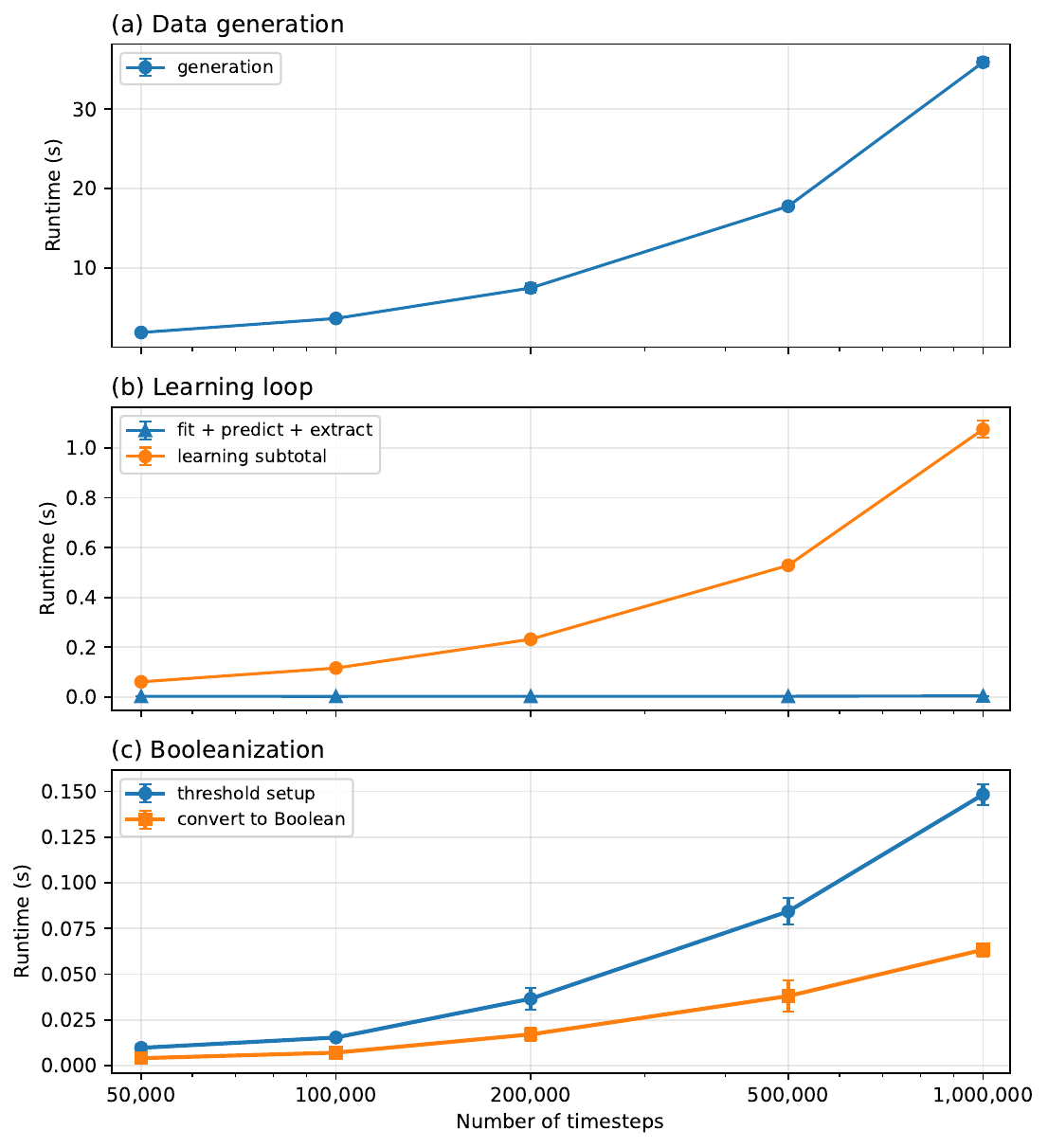}
\caption{Runtime breakdown of the main offline stages versus number of timesteps. (a) Data-generation cost. (b) Learning-loop cost excluding generation, showing the decision-tree fit/predict/extract stage and the subtotal of the non-generation learning stages. (c) Booleanization cost, showing threshold setup and conversion to Boolean.}
\label{fig:runtime_pipelines}
\end{figure}

To place the proposed significance detector in context, we also compare it
with two standard sequential change-detection methods: EWMA, which emphasizes recent observations
through exponentially decaying weights \cite{lucas1990ewma}, and two-sided
CUSUM, which accumulates positive and negative deviations to detect
persistent shifts \cite{page1954cusum}.

All three detectors operate on the same standardized KPI-increment sequences and are calibrated
on held-out control-inactive samples to the same target null false-alarm
rate of $10^{-3}$.  We evaluate representative weak, moderate, and strong
signal regimes corresponding to
$\Delta P_{\mathrm{step}}/\sigma \in \{2,6,10\}$.

\begin{table}[t]
\centering
\caption{Comparison with standard event-detection baselines. All methods
are calibrated to a target null false-alarm rate of $10^{-3}$.}
\label{tab:detection_baselines}
\begin{tabular}{c l c c c}
\hline
$\Delta P_{\mathrm{step}}/\sigma$ & Method & Precision & Recall & F1 \\
\hline
2  & Proposed & 0.527 & 0.133 & \textbf{0.210} \\
   & EWMA     & 0.466 & 0.118 & 0.185 \\
   & CUSUM    & 0.238 & 0.038 & 0.065 \\
\hline
6  & Proposed & 0.764 & 0.987 & \textbf{0.855} \\
   & EWMA     & 0.461 & 0.249 & 0.309 \\
   & CUSUM    & 0.562 & 0.357 & 0.422 \\
\hline
10 & Proposed & 0.730 & 0.995 & \textbf{0.834} \\
   & EWMA     & 0.511 & 0.509 & 0.462 \\
   & CUSUM    & 0.723 & 0.990 & 0.827 \\
\hline
\end{tabular}
\end{table}

Table~\ref{tab:detection_baselines} shows that the proposed detector
provides the strongest performance in the weak and moderate signal regimes
and remains competitive with CUSUM at high signal strength.  The largest
difference occurs at
$\Delta P_{\mathrm{step}}/\sigma=6$, where the proposed method achieves
an F1 score of $0.855$, compared with $0.422$ for CUSUM and $0.309$ for
EWMA.  At the strongest tested signal level, CUSUM nearly matches the
proposed detector, with F1 scores of $0.827$ and $0.834$, respectively.
This behavior is consistent with the event structure of the benchmark:
EWMA and CUSUM accumulate evidence over time, whereas the proposed method
is designed to identify individual control-induced KPI responses directly.

\subsection{Runtime analysis}

Fig.~\ref{fig:runtime_pipelines} reports the computational cost of the main offline stages of the proposed slow loop. Panel~(a) shows the cost of synthetic trace generation. Panel~(b) isolates the downstream slow loop cost excluding generation, separating the decision-tree fit/predict/extract stage from the subtotal of the non-generation learning stages. Panel~(c) reports the corresponding Booleanization cost, separating threshold setup from conversion to Boolean events. These results are intended to assess whether the slow loop analysis remains practical at the offline timescale for which it is designed.

Two trends are clear. First, runtime grows predictably with the number of timesteps in all stages. Second, the dominant cost is trace generation, while the added analysis stages remain comparatively small. The learning subtotal in panel~(b) stays well below the generation cost in panel~(a), and the Booleanization costs in panel~(c) are smaller still. In particular, threshold setup and continuous-to-Boolean conversion both remain lightweight over the tested range relative to the upstream data-generation stage.

The runtime results further support the intended fast/slow separation. In Fig.~\ref{fig:runtime_pipelines}(b), the learning subtotal approaches one second for the largest trace size, already comparable to the upper end of the Near-RT control timescale (\(10\,\mathrm{ms}\)–\(1\,\mathrm{s}\)). Because event detection and model refresh are only part of the full control loop, the dependency representation is more naturally refreshed in a slower loop and then used for faster runtime reasoning.

\subsection{Discussion}

The results suggest a clear overall picture. First, the continuous traces generated by the simulator do contain the latent dependency structure, so the problem is identifiable under chosen settings. Second, the Booleanization step exhibits a clear and interpretable operating-point tradeoff: low thresholds favor sensitivity but admit many spurious detections, whereas moderate-to-high thresholds substantially improve F1 by becoming more selective. Third, increasing signal strength mainly improves recall up to saturation, while the overall threshold trend remains similar across latent dependency graphs of different sizes.

An important implication is that the main bottleneck is not simply raw signal strength alone. In the tested regime, stronger signal quickly removes missed detections, but it does not eliminate a small residual number of false positives. This means that once recall has saturated, further gains are limited less by signal strength than by the detector's false-positive floor and by the calibration of the significance threshold.

Taken together, these results suggest that the Booleanization step can be effective when the significance threshold \(z_{\mathrm{th}}\) is chosen appropriately and the control-induced KPI response is sufficiently separated from background variation. The runtime results also indicate that both the learning-loop and Booleanization computations are practical for offline model refresh, with total cost dominated by data generation rather than by the interpretable inference steps themselves. This supports the intended use of the method within a dynamic/learning architecture, in which heavier continuous-data analysis is confined to the slow loop and downstream reasoning operates on the resulting binary dependency representation.

\section{Conclusion}

This paper studied binary parameter--KPI dependency learning for
AI-RAN control systems, focusing on the event-detection step that
converts noisy continuous telemetry into an interpretable Boolean
representation. To enable controlled evaluation, we developed a
closed-loop synthetic telemetry generator with planted dependencies and
used it to verify that the underlying dependency structure remains
recoverable from the generated telemetry. The results show that
event-detection quality depends primarily on the separation between
control-induced KPI responses and background variation, together with
appropriate calibration of the significance threshold. The proposed
method remains lightweight enough for periodic model refresh in the
slow loop of the intended fast/slow control architecture.

The present evaluation establishes the approach under controlled conditions with state-reactive controllers, enabling precise assessment against known ground truth and providing a foundation for extensions to adaptive controllers and time-varying dependencies.
Future work will extend the framework to adaptive controllers and time-varying dependencies, and will evaluate the approach using real or high-fidelity AI-RAN/O-RAN telemetry as such telemetry becomes available.
 We will also investigate
how the learned Boolean dependency representation can support
downstream conflict detection and mitigation.



\bibliographystyle{IEEEtran}
\bibliography{refs}

\end{document}